\documentclass[11pt,a4paper]{article}
\usepackage[hyperref]{emnlp2018}
\usepackage{hyperref}

\usepackage{times}
\usepackage{latexsym}

\usepackage{url}

\newcommand{\myurl}[1]{\small{\url{#1}}}

\aclfinalcopy %

\title{
Generalizing Word Embeddings using Bag of Subwords
}

\author{Jinman Zhao, Sidharth Mudgal, Yingyu Liang\\
  Department of Computer Sciences \\
  University of Wisconsin-Madison \\
  {\tt \{jz,sidharth,yliang\}@cs.wisc.edu}
}

\date{}

\usepackage{hyperref}

\usepackage{seqsplit}

\newcommand{\BoS}{\text{BoS}}

\usepackage{multirow}

\usepackage{amsmath,amssymb,mathtools}
\newcommand{\R}{\mathbb{R}}

\DeclareMathOperator*{\minimize}{minimize}

\def\1{\mathbbm{1}}
\DeclarePairedDelimiter\abs{\lvert}{\rvert}%
\DeclarePairedDelimiter\norm{\lVert}{\rVert}%

\newcommand{\curlypara}[1]{\left\{{#1}\right\}}
\def\cpara{\curlypara}

\begin{document}
\maketitle
\vspace{-10mm}
\begin{abstract}
  We approach the problem of generalizing pre-trained word embeddings beyond fixed-size vocabularies without using additional contextual information.
  We propose a subword-level word vector generation model that views words as bags of character $n$-grams.
  The model is simple, fast to train and provides good vectors for rare or unseen words.
  Experiments show that our model achieves state-of-the-art performances in English word similarity task and in joint prediction of part-of-speech tag and morphosyntactic attributes in 23 languages,
  suggesting our model's ability in capturing the relationship between words' textual representations and their embeddings.
\end{abstract}

\section{Introduction}

Word embeddings have been an essential part of neural-network based approaches for natural language processing tasks \citep{goldberg2016primer}. However, many popular word embeddings techniques have a fixed vocabulary \cite{word2vec, glove}, i.e., they can only provide vectors over a finite set of common words that appear frequently in a given corpus. Such methods fail to generate vectors for rare words and words not present in the training corpus, but appearing in the test corpus or downstream task texts, raising difficulty for any methods relying on word vectors to efficiently extract useful features from text. This is often referred to as the out-of-vocabulary (OOV) word problem. We aim to address this problem by inferring vectors for OOV words with only access to pre-trained vectors over a fixed vocabulary of common words and the OOV word itself without context.

The motivations come from both linguistics and natural language processing applications. First, from a linguistic view a word can be decomposed into multiple morphemes: stems, affixes, modifiers and etc. This is more often the case for rare words. In some field such as chemistry and agglutinative languages such as Turkish, there exists a systematic way of composing words from morphemes. Some can even be arbitrarily long.

Apart from the explicit and systematic way of making words, we can also observe the ability of a language speaker to infer the meaning of an unseen word. For instance, one can guess that ``pre-EMNLP'' means ``before EMNLP'', even without the presence of any context, suggesting that it is part of our implicit linguistic knowledge to infer meaning of an unseen word solely from its lexical form. This observation, together with the morpheme decomposition of many rare words, implies the feasibility of inferring their vectors from those for common words, and also raises the algorithmic question of how to compute them efficiently.

Second, there are many NLP applications where estimating word embeddings of OOV is critical. For instance, in the case of analyzing Twitter data, while there exists pre-trained word embeddings with giant vocabularies trained on massive number of tweets, such as GloVe vectors \cite{glove}, this would still not cover new words coined by users everyday. In such cases, it is more prudent to extend the available pre-trained vectors trained on very large corpora, so that we can estimate embeddings for OOV words, instead of retraining a new word / subword level embedding model on the new extended data corpus.

OOV words have always been a problem for methods that assume fixed vocabularies. A common workaround is to view all OOV words as a special UNK token and use the same vector for all of them. This would restrict any downstream models from accessing distinct features of those words. Thus, we would like a method to provide vectors that capture semantic and grammatical features even for OOV words. We also would like such method to maximally rely on the word itself, instead of its context, as contextual information is already used later with sentence level models stacking over word vectors.

To achieve this, we aim to build a word embedding model that generalizes pre-trained word embeddings to OOV words.
First, given word embeddings for a fixed vocabulary, our model learns the relationship between the subwords present in each word and its corresponding pre-trained word vector.
Then, using the learned subword information, our model can generate word embeddings for any word, regardless if it is OOV or not.

\paragraph{Contribution}
We propose a simple yet effective subword-level word embedding method that can be efficiently trained given pre-trained word vectors for a limited number of words. Once trained, our embedding model takes the characters $n$-grams in a word as input and gives its word vector as output.\footnote{The code is available at \myurl{https://github.com/jmzhao/bag-of-substring-embedder}.}

Our experiments on word similarity tasks in English and POS tagging in a variety of languages suggests that the proposed word embedder is able to mimic and generalize consistently the word vectors from in-vocabulary words to out-of-vocabulary words, and achieves state-of-the-art scores for the tasks compared to previous subword-level word embedders trained under the same setting. This gives evidence that such a simple model is capable of capturing language speaker's morphological knowledge, and also provides an easy way to generate word vectors for rare or unseen (OOV) words with potential application to various natural language processing tasks.

\paragraph{Related work}
There exist a large body of works that try to incorporate morphological information into word representations, e.g.,~\cite{alexandrescu2006factored,luong2013better,qiu2014co,botha2014compositional,cotterell2015morphological,soricut2015unsupervised}. These approaches typically rely on the morphological decomposition of words. Some other approaches using subword information do not rely on morphological decomposition but requires context information from large text corpus~\cite{schutze1993word,santos2014learning,ling2015finding,wieting2016charagram}.

In particular, \citet{DBLP:journals/tacl/BojanowskiGJM17} introduced fastText, a word embedding method enhanced with subword (character $n$-gram) embeddings. They are able to generate vectors for OOV words, which has been shown useful for text classification~\citep{joulin2016bag}, but the model is to be trained over large text corpus.

\citet{DBLP:conf/emnlp/PinterGE17} use a character-level bidirectional LSTM model called MIMICK, mapping from word strings to word vectors. The idea of using character-level recurrent neural networks (RNNs) for word vectors is not new~\citep{ling2015finding,plank2016multilingual}, but as per authors' knowledge, they are by far the only attempt to the exact task of generalizing word vectors from only pre-trained vectors with a fixed vocabulary, i.e. with no access to contextual information.

\section{Bag-of-Substring Model}
\newcommand{\subs}[2]{subs_{#1}^{#2}}
\newcommand{\<}{\texttt{<}}
\renewcommand{\>}{\texttt{>}}

Our Bag-of-Substring (\BoS) word vector generation model views a word as a bag of its substrings, or character $n$-grams. Specifically, we maintain a vector lookup table for each possible substrings (or character $n$-grams) of length between $l_{min}$ and $l_{max}$. A word vector is then formed as the average of vectors of all its substrings with lengths in the range.
Let $\Sigma$ be the finite set of characters in the language, $\subs{a}{b}(s) = \cpara{t \text{ is substring of } s : a \leq \abs{t} \leq b}$ for string $s \in \Sigma^*$ be the set of substrings of $s$ whose length is between $a$ and $b$ inclusive, and $\<s\>$ be the concatenation of character \<, string $s$ and character \> where $\<, \> \not\in \Sigma$.
The BoS embedding for a string/word $s$ can be expressed as
\begin{equation} \label{eqn:bos}
  \BoS(s; V) = \frac{1}{\abs{\mathcal{S}_{\<s\>}}} \sum_{t \in \mathcal{S}_{\<s\>}} v_{t},
\end{equation}
where $V \in \R^{d \times (\abs{\Sigma}^{l_{min}} + \dots + \abs{\Sigma}^{l_{max}})}$ are the parameters which
stores the embeddings of dimension $d$ for each possible substring of length between $l_{min}$ and $l_{max}$, $v_t$ is the vector in $V$ indexed by $t$, $\mathcal{S}_{\<s\>}$ is a shorthand for $\subs{l_{min}}{l_{max}}(\<s\>)$.
Special characters $\<, \> \not\in \Sigma$ are used to mark the start and the end of the word and thus help the model to distinguish homographic morphemes that occur at different word parts, e.g. prefixes or suffixes.
An example BoS representation for word \texttt{infix} is
$\subs{3}{4}(\<\texttt{infix}\>)=\{$\texttt{<in, <inf, inf, infi, nfi, nfix, fix, fix>, ix>}$\}$.

fastText~\citep{DBLP:journals/tacl/BojanowskiGJM17} uses the same idea for their word vector generation part.
However, unlike them, we train the model directly towards pre-trained vectors, instead of via context prediction over text corpora.

\paragraph{Training}
Given pre-trained vectors for a set of common words, our model views them as targets and is trained to fit these targets. Once the parameters (the vectors $v_t$ for the substrings) are learned, the model can then be used to infer vectors for rare words.
Let $U \in \R^{d \times \abs{W}}$ be the target vectors of the same dimension $d$ over finite vocabulary $W \subset \Sigma^*$.
Our model is trained by minimizing the overall loss between the generated and the given vectors for each word:
\begin{equation}
  \minimize_{V} \frac{1}{\abs{W}} \sum_{w \in W} l(BoS(w; V), u_w)
\end{equation}
where the loss function $l(v, u) = \frac{1}{2} \norm{v - u}_2^2$, namely the mean squared loss.

After training, one can use the learned $V$ and Eqn~(\ref{eqn:bos}) to compute the vector for any given word, even if it is OOV. 

\paragraph{Hyperparameters}
We set the following hyperparameters for all the experiments. For BoS model, $l_{min}=3$ and $l_{max}=6$ following \citet{DBLP:journals/tacl/BojanowskiGJM17}. Note that under this setting, $\mathcal{S}_{\<s\>}$ can never be empty for non-empty string $s$. For optimization, stochastic gradient descent with learning rate 1 for 100 epochs. The dimension of the word vectors is not a hyperparameters here as it needs to agree with the target vector.

\section{Word Similarity}\label{sec:intrinsic}
We run experiments to quantitatively evaluate the our model's generalizability towards OOV words.

The word similarity task asks to predict word similarity between a pair of two words. Given a set of pairs of words and gold labels for their similarities, the performance of word embeddings is measured by the correlation between the gold similarities and the similarities induced by the generated embeddings. And we can thus imply how good our model is at generating word vectors. The word similarity here is computed using the cosine distance between the two word vectors, and the correlation is computed using Spearman's $\rho$.

\paragraph{Datasets}
We evaluate over Stanford RareWord (RW) introduced by \citet{DBLP:conf/conll/LuongSM13} and WordSim353 (WS) introduced by \citet{finkelstein2001placing}.
RW consists of less common words so we use it to access our model's ability to generalize word embeddings to OOV words.
WS is composed of mostly common words and we use it to test if our subword-level models successfully mimic the target vectors.

\paragraph{Target vectors}
We train our BoS model over the English Polyglot vectors~\footnote{\myurl{http://polyglot.readthedocs.io/en/latest/Download.html}} to establish a direct comparison with results from MIMICK~\citep{DBLP:conf/emnlp/PinterGE17},
and as well as the Google word2vec vectors~\footnote{\myurl{https://code.google.com/archive/p/word2vec/}}
which are popularly used in NLP tasks.
Polyglot~\citep{polyglot:2013:ACL-CoNLL} is a multilingual NLP dataset, which also provides pre-trained word vectors over each language's corpus with a vocabulary of 100,000 most frequent words.
For Google vectors, most of their vocabulary consists of non-words such as URLs and phrases, so we normalize tokens into ASCII characters by taking off all the diacritics and take only tokens consisting of a single word with all lower letters.
Statistics of the processed vectors are summarized in Table~\ref{tab:target-stats}, along with their word similarity task scores (for in-vocabulary words only) and OOV rate over the aforementioned evaluation sets.

\begin{table}
  \small
  \centering
  \begin{tabular}{c | c c | c c}
    \hline\hline
      & Dim. & \# Tokens & RW & WS \\
    \hline
    Polyglot & 64 & 100k & 41(58\%) & 45(5\%) \\
    Google & 300 & 160k & 53(11\%) & 69(1\%) \\
    \hline\hline
  \end{tabular}
  \caption{\small Target vectors statistics and word similarity task scores in Spearman's $\rho \times 100$.
  In parentheses are OOV rates.
  }
  \label{tab:target-stats}
\end{table}

\begin{table}
  \small
  \centering
  \begin{tabular}{l | c c | l l}
    \hline\hline
    Model        & Size  & Target   & RW & WS \\
    \hline
    EditDist     & -     & -        & 18 & -2 \\
    \hline
    MIMICK       & 649KB     & Polyglot & 14 & 12 \\
    $\BoS$   & 238MB & Polyglot & 36 & 36 \\
    $\BoS$   & 1.3GB & Google   & {46} & {56} \\
    \hline
    $\text{fastText}$ & 8.0GB & - & 48 & 74 \\
    \hline\hline
  \end{tabular}
  \caption{\small
  Word similarity task results measured in Spearman's $\rho \times 100$.
  }
  \label{tab:wordsim-results}
\end{table}

\paragraph{Baselines}
We compare the scores with other subword-level models (fastText and MIMICK) and word similarity induced by non-parametric edit distance (EditDist).

fastText~\cite{DBLP:journals/tacl/BojanowskiGJM17} uses the same subword-level character $n$-gram model but is to be trained via context prediction over large text corpora (here English Wikipedia dump~\footnote{\myurl{https://fasttext.cc/docs/en/pretrained-vectors.html}}).
MIMICK~\cite{DBLP:conf/emnlp/PinterGE17} is a character-level bidirectional LSTM word embedder trained against pre-trained word vectors (here Polyglot vectors~\footnote{\myurl{https://github.com/yuvalpinter/Mimick}}).

Edit distance is defined between two strings as the smallest number of modifications: adding, deleting and changing one character, needed to turn one string into the other. It can be computed using dynamic programming in $O(\abs{s_1} \times \abs{s_2})$ time.
The word similarity between $w_1$ and $w_2$ here is the edit distance normalized by the length of the longer word:
\begin{equation}
  s_{EditDist}(w_1, w_2) = - \frac{d_{edit}(w_1, w_2)}{\max(\abs{w_1}, \abs{w_2})}
\end{equation}
where $d_{edit}$ is edit distance.

\paragraph{Results}
Results are summarized in Table~\ref{tab:wordsim-results}.
When trained over Polyglot vectors, our BoS model works better than EditDist and MIMICK.
When trained on Google vectors,
the correlation scores are almost as good as those of fastText,
the state-of-the-art subword level word embedder.
However, unlike fastText, our model does not have access to word contexts in a large text corpus for training.
In both cases, the significant differences of scores compared to those of EditDist,
suggest that our model indeed learns to capture semantic similarities between words, rather than superficial similarities in spelling.

Comparing to MIMICK, our model is able to fill up 81\% (14 to 36 against 41) and 73\% (12 to 36 against 45) of the gaps in scores over RW and WS respectively.
This improvement is more significant on RW with most (58\%) of its words are OOV for the PloyGlot vectors, suggesting our model's power in generating consistent word vectors for OOV words.
Surprisingly MIMICK performs no better than the edit distance baseline when evaluated on RW.
Combined with the fact that it does no better for WS which has a near-zero OOV rate,
it suggests MIMICK's limited power of generalizing word vectors towards OOV words,
or even reproduce consistent word vector for in-vocabulary words.
As a sanity check, we see that all of the embedder models scores obviously better than EditDist when evaluated over common words (WS),
showing that all of them are able to at least remember or mimic the word vectors for in-vocabulary words.

Also note that our model is fast to train. With a naive single-thread CPU-only Python implementation, it can finish 100 epochs of training over English PolyGlot vectors within 352 seconds on a machine with an Intel Core i7-6700 (3.4 GHz) CPU, 32GB memory and 1TB SSD. Compared to fastText which, with a fast multithread C++ implementation, takes hours to be trained over giga bytes of text corpus, our method provides a cheap way to generalize reasonably good word vectors for OOV words.

\begin{table*}[!t]
  \small
  \centering
  \begin{tabular}{l | r | c c c | c c c}
    \hline\hline
  & \multirow{2}{*}{$N_{train}$} & \multicolumn{3}{c}{POS tagging} & \multicolumn{3}{c}{Morphosyntactic attributes} \\
  &  & random & MIMKCK & BoS & random & MIMICK & BoS \\
  \hline
Kazakh & 4,949 & 0.589 & 0.681 & \textbf{0.758}(0.077) & 0.021 & 0.032 & \textbf{0.240}(0.208) \\
Tamil & 6,329 & 0.480 & 0.678 & \textbf{0.774}(0.097) & 0.568 & 0.673 & \textbf{0.762}(0.089) \\
Latvian & 13,781 & 0.589 & 0.757 & \textbf{0.872}(0.115) & 0.374 & 0.572 & \textbf{0.676}(0.104) \\
Vietnamese & 31,800 & 0.749 & 0.564 & \textbf{0.846}(0.282) & - & - & - \\
Hungarian & 33,017 & 0.594 & 0.858 & \textbf{0.922}(0.065) & 0.569 & 0.775 & \textbf{0.836}(0.061) \\
Turkish & 41,748 & 0.636 & 0.767 & \textbf{0.890}(0.123) & 0.543 & 0.776 & \textbf{0.826}(0.050) \\
Greek & 47,449 & 0.819 & 0.907 & \textbf{0.965}(0.058) & 0.783 & 0.903 & \textbf{0.934}(0.031) \\
Bulgarian & 50,000 & 0.804 & 0.903 & \textbf{0.971}(0.068) & 0.649 & 0.851 & \textbf{0.915}(0.064) \\
Swedish & 66,645 & 0.748 & 0.813 & \textbf{0.945}(0.132) & 0.707 & 0.812 & \textbf{0.930}(0.118) \\
Basque & 72,974 & 0.662 & 0.823 & \textbf{0.913}(0.091) & 0.564 & 0.778 & \textbf{0.820}(0.042) \\
Russian & 79,772 & 0.665 & 0.897 & \textbf{0.948}(0.051) & 0.592 & 0.855 & \textbf{0.915}(0.060) \\
Danish & 88,980 & 0.788 & 0.834 & \textbf{0.947}(0.114) & 0.745 & 0.813 & \textbf{0.927}(0.114) \\
Indonesian & 97,531 & 0.724 & 0.788 & \textbf{0.915}(0.127) & - & - & - \\
Chinese & 98,608 & 0.721 & 0.793 & \textbf{0.835}(0.042) & 0.699 & 0.767 & \textbf{0.790}(0.022) \\
Persian & 121,064 & 0.843 & 0.866 & \textbf{0.957}(0.091) & 0.745 & 0.792 & \textbf{0.918}(0.125) \\
Hebrew & 135,496 & 0.814 & 0.858 & \textbf{0.957}(0.099) & 0.648 & 0.837 & \textbf{0.903}(0.066) \\
Romanian & 163,262 & 0.796 & 0.874 & \textbf{0.956}(0.082) & 0.718 & 0.876 & \textbf{0.942}(0.066) \\
English & 204,587 & 0.770 & 0.826 & \textbf{0.932}(0.106) & 0.822 & 0.859 & \textbf{0.947}(0.089) \\
Arabic & 225,853 & 0.780 & 0.901 & \textbf{0.950}(0.049) & 0.711 & 0.901 & \textbf{0.942}(0.041) \\
Hindi & 281,057 & 0.824 & 0.848 & \textbf{0.939}(0.091) & 0.863 & 0.888 & \textbf{0.951}(0.063) \\
Italian & 289,440 & 0.810 & 0.909 & \textbf{0.964}(0.056) & 0.839 & 0.927 & \textbf{0.964}(0.037) \\
Spanish & 382,436 & 0.819 & 0.914 & \textbf{0.959}(0.045) & 0.793 & 0.915 & \textbf{0.954}(0.038) \\
Czech & 1,173,282 & 0.695 & 0.908 & \textbf{0.966}(0.058) & 0.622 & 0.845 & \textbf{0.905}(0.061) \\
  \hline\hline
  \end{tabular}
  \caption{\centering 
  POS tagging accuracy and morphosyntactic attributes micro F1 over 23 languages (UD 1.4). 
  In parentheses are the gains to MIMICK.}
  \label{tab:pos-tagging-results}
\end{table*}

\section{Joint Prediction of Part-of-Speech Tags and Morphosyntactic Attributes}
Besides word similarity, we try to access our embedders' ability of capturing words' syntactic and semantic features by evaluating with the task of predicting part-of-speech (POS) tags and morphosyntactic attributes for words in a sentence.
For each word in a given sentence, the task asks for a POS tag and a label for each applicable morphosyntactic category, such as gender, case or tense.

\paragraph{Dataset}
We use Universal Dependencies (UD) dataset~\cite{petrov2012universal} for this task.
UD is an open-community effort to build consistent annotated treebank cross many languages.
We pick the specific version 1.4 to enable a direct comparison with \citet{DBLP:conf/emnlp/PinterGE17}.
Since we use PolyGlot vectors to train our word embedders, we conduct experiments on the 23 languages that appear in both Polyglot and UD 1.4.

\paragraph{Model}
We adopt the same sentence-level bidirectional LSTM model from \citet{DBLP:conf/emnlp/PinterGE17} for the joint prediction of both labels.
Given a sentence as a sequence of words, we first embed each word using the word embedder we choose and then fed the embeddings into the LSTM.
The output of LSTM is then used to predict POS and morphosyntactic tags.

We emphasize the difference in the setting that
we \emph{fix} the word embeddings during the training,
as to better evaluate the ability and consistency of the embeddings in capturing words' semantics and syntactics,
rather than LSTM's ability to memorize words and infer the role of words from their context.

We use the same set of hyperparameters for the LSTM model as Sec. 5.3 in \citet{DBLP:conf/emnlp/PinterGE17} and train the model for 20 epochs for each language.
The BoS and MIMICK word embedders are trained beforehand with PolyGlot dataset using the same way described earlier.

\paragraph{Results}
The POS tagging accuracies and micro F1 scores for morphosyntactic attributes are reported in Table~\ref{tab:pos-tagging-results} with word vectors generated by different models.
The BoS and MIMICK model here are trained against Polyglot vectors.
As a comparison, we include the results using random word vectors of the same dimension (64).

Our BoS model shows steady and significant gain compared to MIMICK embeddings for both tasks in \emph{all} languages.
We especially observe the greatest margins for agglutinative languages such as Turkish and Indonesian, and in Germanic languages English, Swedish and Danish,
suggesting that our model learns stable representations for morphemes to consistent word type signal.

\section{Conclusion}
We proposed a subword-level word embedding model and a word vector generalization method that enables extending pre-trained word embeddings with fixed size vocabularies to estimate word embeddings for out-of-vocabulary words. Intrinsic evaluation on word similarity tasks and extrinsic evaluation on POS tagging task demonstrate that our model captures morphological knowledge and generates good estimates of word vectors for OOV words.

\section*{Acknowledgements}
We would like to thank the anonymous reviewers for helpful comments. This work was supported in part by FA9550-18-1-0166. Y. L. would also like to acknowledge that support for this research was provided by the Office of the Vice Chancellor for Research and Graduate Education at the University of Wisconsin-Madison with funding from the Wisconsin Alumni Research Foundation.

\bibliography{main}
\bibliographystyle{acl_natbib_nourl}

\end{document}